\documentclass[letterpaper, 10 pt, conference]{ieeeconf}
\usepackage{graphicx}
\usepackage{amsmath}
\usepackage{longtable}
\usepackage{float}
\usepackage{epstopdf}
\newcommand{\tseg}{{\tt OffRoadTranSeg }}
\newcommand{\re}{{\tt RELLIS-3D }}
\newcommand{\ru}{{\tt RUGD }}
\IEEEoverridecommandlockouts 
\title{OffRoadTranSeg: Semi-Supervised Segmentation using Transformers on OffRoad environments}

\author{
   Anukriti Singh$^1$,
   
   Kartikeya Singh$^1$, and P.B. Sujit %
   \thanks{The authors are with the IISER Bhopal. $^*$ Equal contribution.}
}
\date{}

\begin{document}
\maketitle
\begin{abstract}
We present OffRoadTranSeg, the first end-to-end framework for semi-supervised segmentation in unstructured outdoor environment using transformers and automatic data selection for labelling. The offroad segmentation is a scene understanding approach that  is widely used in autonomous driving. The popular offroad segmentation method is to use fully connected convolution layers and large labelled data, however, due to  class imbalance, there will be several mismatches and also some classes may not be detected. Our approach is to do the task of offroad segmentation in a semi-supervised manner. The aim is to provide a model where self supervised vision transformer is used to fine-tune offroad datasets with self-supervised data collection for labelling using depth estimation. The proposed method is validated on \re and \ru  offroad datasets. The experiments show that \tseg outperformed other state of the art models, and also solves the \re class imbalance problem.  
\end{abstract}



\section{Introduction}
	
An autonomous outdoor navigation \cite{Cordts2016Cityscapes, alhaija2018augmented}  system for unstructured environment \cite{procopio2009learning} can be built using semantic segmentation network. It has applications in areas like inspection, exploration, rescue, and reconnaissance. But scene semantic segmentation in off-road environments is challenging due to absence of a defined boundaries between classes and terrain structure. This leads to challenges like class imbalance when algorithms are deployed in real environments. However, class imbalance problem is highly environment specific and becomes more severe when its within a dataset, like in RELLIS-3D \cite{jiang2020rellis}. This makes semantic segmentation in off-road classes challenging as compared to urban environments. Presently, Convolutional Neural Networks (CNNs) \cite{726791} in semantic segmentation \cite{7298965} have achieved state-of-the-art results for unstructured outdoor environments like RELLIS-3D. However, CNNs require dense annotated datasets, which is costly and takes time specially in off-road environments where classes that are only visible in few frames can be important for autonomous decision making and navigation planning. 

CNNs typically fail to produce accurate masks when the content of the images are complicated. For example in \re off-road dataset Jiang et\:al. \cite{jiang2020rellis} used HRNet for semantic segmentation. They successfully segmented classes like sky, grass, tree, etc. but there were some exceptions like log, water, and building where the mIOU was 0. Guan et\:al. \cite{guan2021ganav} and Viswanath et\:al. \cite{viswanath2021offseg} performed segmentation after classifying groups of terrain classes based on
traversal regions. \cite{guan2021ganav} used group attention network with transformers for segmentation and \cite{viswanath2021offseg} used color clustering. However, since both of them pooled classes they did not have to face the challenge of class imbalance. Recently, it was proposed to use transformers for vision based tasks \cite{dosovitskiy2020image} like semantic segmentation \cite{strudel2021segmenter, zheng2020rethinking, zhu2020deformable} after their success in Natural Language Processing(NLP).  Jin et\:al. \cite{jin2021trseg}  showed how transformers based model can perform better for semantic segmentation on two datasets: RUGD and Cityscape. Although, their method was completely supervised and required all the images in the dataset to be annotated. Caron et\:al. \cite{caron2021emerging}  proposed DINO to use self supervised pretraining on vision transformers, which can benefit segmentation as well as classification tasks in vision. DINO is pretrained on ImageNet dataset and uses knowledge distillation \cite{hinton2015distilling}. 

\begin{figure*}[h!]
    \centering
    \includegraphics[width=14cm, height=5cm]{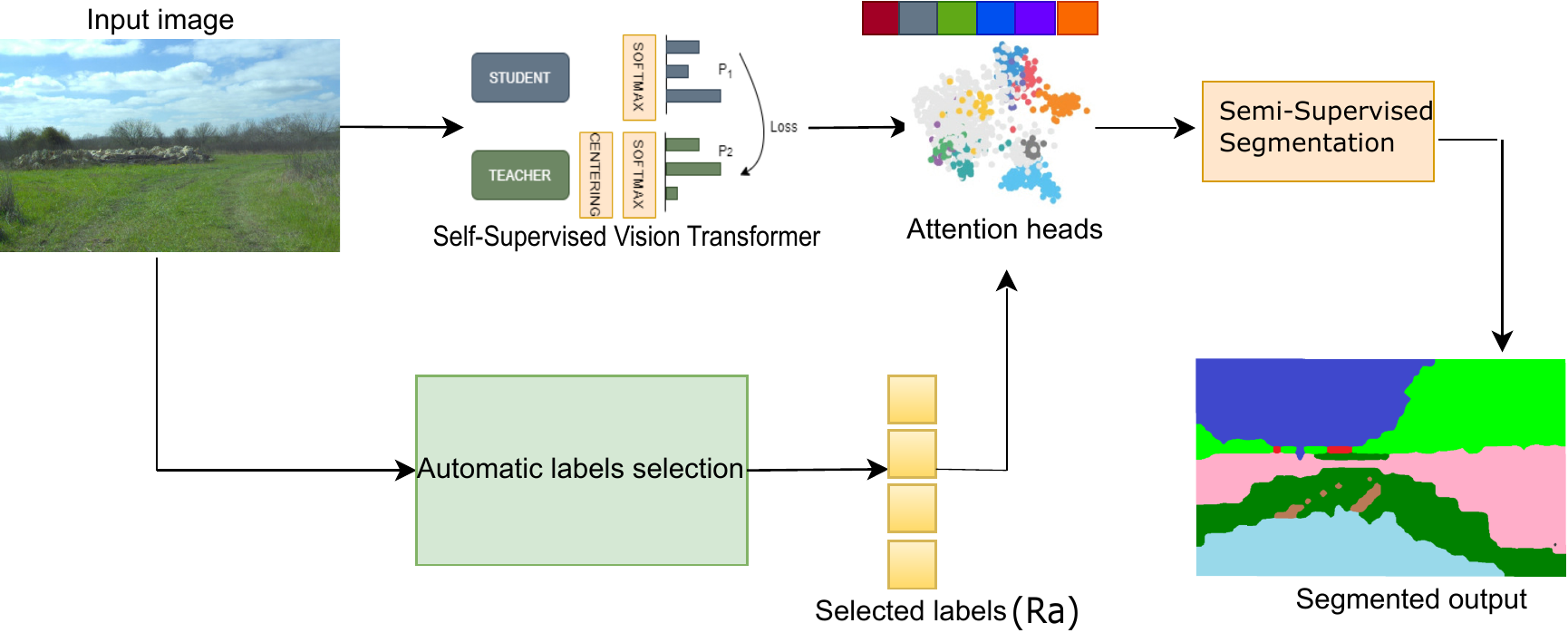}
    \caption{Our framework consists of two stages: Fine-tuning self supervised vision transformer to get the attention maps and automatically selecting labelled images for evaluating the network for semi supervised semantic segmentation}
    \label{fig:labelscompare}
\end{figure*}

\subsection*{Main Results:}  Annotations for off road datasets is particularly challenging since the boundaries are mixed and not well defined, it takes a lot of time and increases stress leading to mislabelling. In this paper, we present a novel semi-supervised approach for offroad semantic segmentation. The approach involves fine-tuning the DINO network with no labelled data and evaluating it with the automatic selected data. Our method is general and is particularly designed for unstructured outdoor environments. Semi-supervised sequence segmentation requires to provide the mask of the first image in the sequence and the network segments the subsequent frames based on that. We instead propose to do an automatic data selection, where the selected images are labelled instead of just the first frame. In summary, the main contributions of our proposed method are : 
\begin{itemize}
\item A novel model for semi-supervised semantic segmentation in unstructured outdoor environment using self attention in vision transformers.

\item Completely solving class imbalance problem by obtaining mIoU value for all the classes in \re dataset. We also test our model on other offroad datasets like \ru and obtain IoU values for multiple classes better than those reported in \ru \cite{wigness2019rugd}.  

\item Significantly reducing the amount of labelled data required for training and evaluating using an algorithm to automatically select the most diverse set of images per sequence. We used less than 25 annotated images during training.
\end{itemize}

We evaluate our method on \re \cite{jiang2020rellis} which contains 20 classes and 13, 556 images and \ru \cite{wigness2019rugd} dataset with 24 classes and over 29, 000 images. \tseg outperforms the networks used in \cite{jiang2020rellis} and \cite{wigness2019rugd} in the semi supervised fashion. 

\begin{figure*}[h!]
    \centering
    \includegraphics[width=13.5cm, height=7cm]{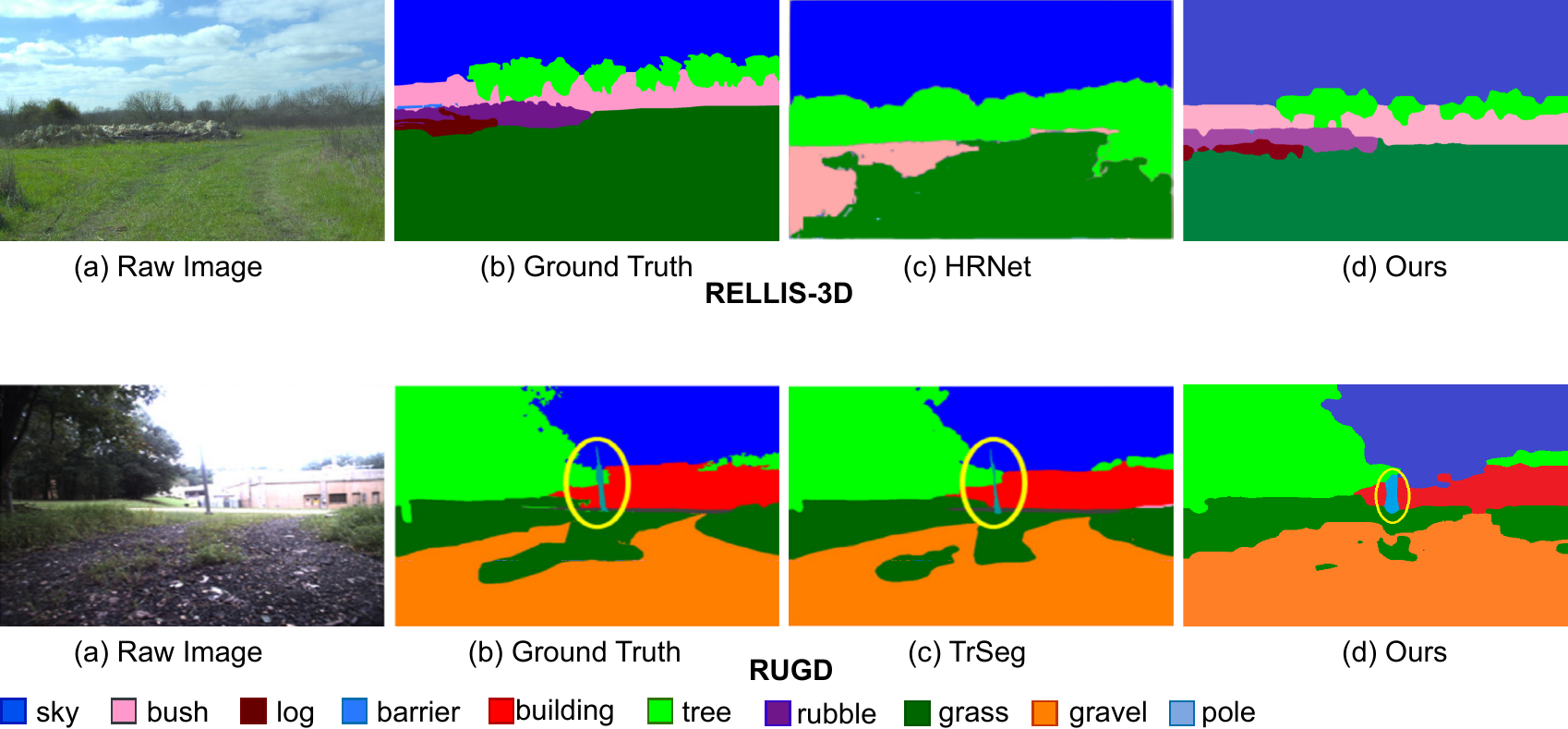}
    \caption{Both the columns highlight the raw image, ground truth, performance of SOTA algorithm and our approach on \re and \ru dataset. We observe that \tseg is able to segment all the classes of the image as present in the raw image.}
    \label{fig:labelscompare}
\end{figure*}



\section{Related Work}
\subsection{Semi-supervised semantic segmentation}
Segmentation is a widely studied area using convolutional neural network as well as the present growing transformer networks. The semantic segmentation tasks can be supervised, semi supervised and self supervised. A number of architectures have been proposed to perform segmentation, like PSPnet \cite{zhao2017pyramid}, HRNETV2+OCR \cite{wang2020deep}, BiSeNetV2 \cite{yu2020bisenet} in CNNs and DETR \cite{zhu2020deformable}, SETR \cite{zheng2020rethinking} in transformer based networks. However, the CNNs based network performs well in urban navigation where the roads and structure is defined, they lack in performance in offroad unstructured areas due to undefined, overlapping boundaries and convoluted area.  
\subsection{Transformers}
Transformers were initially proposed for their use in Natural Language Processing (NLP). With their increasing good performance, transformers were then introduced in computer vision with CNNs, making a hybrid model in DETR \cite{zhu2020deformable}. Later ViT \cite{dosovitskiy2020image} was proposed which proved to be a successful transformer network for vision tasks like object classification. Transformer based network have benefits over purely CNN architectures since they have high accuracy with less computation time for training. TrSeg\cite{JIN202129} used transformer based network for segmentation in urban area dataset like cityscape and offroad dataset \ru, however their network performs supervised segmentation. DINO is a self supervised network which predicts attention in the images and can also be used in semantic segmentation. Our network is based on DINO, which makes segmentation in unstructured environments semi supervised. It has explicit information about the semantic segmentation of an image, which does not emerge as clearly with supervised ViTs, nor with convnets.
\subsection{Dataset}
Navigation is an important part in perception module of  urban and unstructured offroad regions. Latest datasets on offroad are \re and \ru which are  especially tailored for semantic segmentation in a variety of natural, unstructured semi-urban areas. \re is the latest dataset which is hugely derived from \ru and has additionally added unique terrain classes which aren't present in \ru. Both these datasets have rich ontology and large annotations of their ground truths. \re dataset also includes 3-D lidar data and their annotations. We use both these latest datasets to evaluate the performance of our proposed framework.   

\begin{table*}
\centering
\resizebox{2\columnwidth}{!}{
\begin{tabular}{|l|l|l|l|l|l|l|l|l|l|l|l|l|l|l|l|l|}
 \hline
 \multicolumn{17}{|c|}{\textbf{IoU Distribution for $\textbf{RELLIS-3D}$ on \tseg}} \\
 \hline
 \textbf{Sky}&\textbf{Bush}&\textbf{Building}&\textbf{Log}&  \textbf{Grass}&\textbf{Person}&\textbf{Tree}&\textbf{Asphalt}&\textbf{Rubble} & \textbf{Mud}&\textbf{Fence}&\textbf{Puddle}&\textbf{Concrete}&\textbf{Barrier}&  \textbf{Vehicle}&\textbf{Object}&\textbf{Pole}\\
 \hline
 97\%  &96\%   &92\%&96\%&92\%&40\%&82\%&97\%&77\%&97\%&88\%&92\%&98\%&90\%&98\%&77\%&87\%\\
 \hline
 \multicolumn{17}{|c|}{\textbf{IoU Distribution for $\textbf{RELLIS-3D}$ on HRNet}} \\
 \hline
 98\%  &78\%   &1\%  &0\%& 93\%&89\%&91\%&67\%&77\%&  96\%&88\%&64\%&92\%&67\%&42\%&50\%&49\%\\\hline
\end{tabular}%
}
\caption{Class-wise IoU distribution for \re dataset on \tseg and HRNet.}
\label{table:rellis}
\end{table*}

\begin{table*}
\centering
\resizebox{2\columnwidth}{!}{ 
\begin{tabular}{|l|l|l|l|l|l|l|l|l|l|l|l|l|l|l|l|l|l|l|l|l|}
 \hline
Class& \textbf{Dirt}&\textbf{Rockbed}&\textbf{Sand}&\textbf{Grass}&\textbf{Log}&\textbf{Bicycle}&  \textbf{Tree}&\textbf{Mulch}&\textbf{Pole}&\textbf{Fence}& \textbf{Bush}&\textbf{Sky}&\textbf{Sign}&\textbf{Vehicle}&\textbf{Rock}&\textbf{Concrete}&  \textbf{Bridge}&\textbf{Asphalt}&\textbf{Gravel}&\textbf{building}\\
 \hline
 IoU& 88\%  &94\%   &92\%   &97\%&89\%&91\%&82\%&86\%&94\%&90\%& 76\%&96\%&88\%&72\%&86\%&88\%&91\%&96\%&89\%&93\%\\
 \hline
\end{tabular} %
}
\caption{Class-wise IoU distribution for \ru dataset on \tseg. }
\label{table:rugd1}
\end{table*}

\begin{table*}
\centering
\resizebox{2\columnwidth}{!}{
\begin{tabular}{|l|l|l|l|l|l|l|l|l|l|l|l|l|l|l|l|l|l|l|l|l|}
 \hline
Class& \textbf{Dirt}&\textbf{Rockbed}&\textbf{Sand}&\textbf{Grass}&\textbf{Log}&\textbf{Bicycle}&  \textbf{Tree}&\textbf{Mulch}&\textbf{Pole}&\textbf{Fence}& \textbf{Bush}&\textbf{Sky}&\textbf{Sign}&\textbf{Vehicle}&\textbf{Rock}&\textbf{Concrete}&  \textbf{Bridge}&\textbf{Asphalt}&\textbf{Gravel}&\textbf{building}\\
 \hline
 IoU& 0\%  &10\%   &27\%   &77\%&10\%&-&83\%&45\%&19\%&45\%& 
 27\%&81\%&11\%&60\%&9\%&83\%&0\%&13\%&38\%&73\%\\
 \hline
\end{tabular}%
}
\caption{Class-wise IoU distribution for \ru dataset taken from their benchmark method 4(ResNet50+UperNet).}
\label{table:rugd2}
\end{table*}

\section{OffRoadTranSeg}
The proposed \tseg architecture is illustrated in Fig. 2 and it contains two major component : Self-Supervised Transformer and Automatic Data Selection. With \tseg off-terrain unstructured environment is semantically segmented using self supervised attention maps and automatic data selection. Our framework is motivated by DINO\cite{caron2021emerging} a state of the art self supervised attention mechanism based on student teacher transformer network which performed well on ImageNet dataset and DAVIS-2017 video instance segmentation benchmark \cite{ponttuset20182017}. 
This framework takes the advantage of attention in transformers and addresses the class imbalance problem which was introduced with \re dataset in semantic segmentation of offroad images. 
The details of our \tseg model for semi supervised semantic segmentation will be described in the following subsections. 

\subsection{Transformer}
\tseg is based on a fully transformer-based architecture, partially borrowed from knowledge distillation\cite{hinton2015distilling} and self-supervised approaches used in DINO. Knowledge distillation has two network: student network and a teacher network. The student network $S_{\Theta_{s}}$ is trained to match the output of a given teacher network $S_{\Theta_{t}}$, parameterised by $\Theta_{s}$ and $\Theta_{t}$ respectively. When given an input image $x$, probability distribution $P_{s}$ and $P_{t}$ from both the networks are calculated as an output over the dimension $K$. The output of the network, probability $P$ is normalized using a softmax function described below:
\begin{equation}
P_{s}(x)^{(i)} = \frac{\exp(S_{\Theta_{s}}(x)^{(i)}/T_{s})}{\sum_{k=1}^{K}\exp(S_{\Theta_{s}}(x)^{(k)}/T_{s})} 
\end{equation}
Where the temperature parameter $T_{s} > 0$ controls the sharpness of the output distribution. The similar equation holds true for probability distribution $P_{t}$. In DINO both the student and the teacher architecture $S$ share the same network with different set of parameters, $\Theta_{s}$ and $\Theta_{t}$. This self supervised vision transformer network has two types of backbone architectures $f$, ViT\cite{dosovitskiy2020image} and ResNet-50\cite{DBLP:journals/corr/HeZRS15} and a projection head $h$. For unstructured terrain image dataset we use both these networks as backbone. The projection head $h: g = h ◦ f$, which works best for this network consists of a three layer multi-layer perceptron with 2048 hidden dimension followed by $l_{2}$ normalization and a weight
normalized fully connected layer. Also, The DINO network does not have batch normalizations (BN) by default.

\subsection{Automatic Data selection (ADS)}
Self-supervised monocular depth estimation(SDE) \cite{hoyer2020three} a type of label selection method which makes the process of selecting the image to annotate automatic. Recently, manually labelling data is starting to get replaced with self supervised learning. Segmentation is tightly connected with depth estimation. The leverage of this is taken to improve semantic segmentation in unstructured environments with self supervised depth estimation. Labelling images in offroad is particularly hard because it requires fine grained segmentation. With this approach we come up with a way which requires comparatively less number of annotated images and improve the performance of segmentation. 
Self supervised depth estimation is used as a pretext task for our offroad data, which has $N$ number of total samples and $N_{s}$ are selected for annotation. Here we denote, $R$ as total number images in the data, $R_{A}$ as the subset of images selected for annotation and $R_{U}$ as the un-selected images subset. Initially, all data is un-selected and hence, $R=R_{U}$ and selected images for annotation is empty. For the selection of images we first do diversity sampling, to make sure most diverse set of images are selected and which cover different scenes in off-trail image datasets. The set of images choosen for annotation should be diverse enough so that they represent the entire dataset. The intermediate layer of SDE network calculates features $\Phi_{j}^{SDE}$ over which we calculate L2 distance and do iterative farthest point sampling. At step t, for each of the samples at that point, we choose the image from $R_{U}$ which has the largest distance to the current annotation set $R_{A}$. The selected sample of images $R_{A}$ is iteratively expanded by moving one image at
a time from $R_{U}$ to $R_{A}$ until the required number of images are collected:\\ 
\begin{equation}
i = \underset{I_{i}\epsilon R_{U}}{\arg\max}\underset{I_{i}\epsilon R_{A}}{\min}
\left \| \Phi_{i}^{SDE}-\Phi_{j}^{SDE}\right \|
\\
\end{equation}
   
The next selection criteria is uncertainty sampling. When the model is trained on $R_{A}$, the images with high uncertainties near the decision boundary are favoured.For uncertainty sampling, we need to train and update the
model with $R_{A}$. It is inefficient to repeat this every time a new image is added. To make it more efficient, 
the selection is divided into T steps and the model is trained T times.
In each step t, the required number of images are selected and moved from unselected subset to selected images for annotation. With uncertainty Sampling the aim is to select difficult samples, i.e., samples in $R_{U}$ that the model trained on the current $R_{A}$ cannot handle well since the dataset of offroad is very diverse and uncertain. The equation is then updated : 
\begin{equation}
i = \underset{I_{i}\epsilon R_{U}}{\arg\max}\underset{I_{i}\epsilon R_{A}}{\min}
\left \| \Phi_{i}^{SDE}-\Phi_{j}^{SDE}\right \| - \lambda _{E}E(i) 
\end{equation}
Using automatic data selection we get a set of selected annotated data. To perform semi-supervised sequence segmentation a full mask of the object(s) of interest in the first frame of a video sequence has to be given and the network produce the segmentation mask for that object(s) in the subsequent frames. We leverage automatic data selection in semi supervised video segmentation, where instead of giving annotation of only the first frame we select the most diversed frames $n_{t}$ from a sequence and divide it in batches.

\section{Experiments and Results}\label{sec:result}
In this section we provide the implementation details of \tseg and describe the experiments performed to highlight the two main components of our approach.

\begin{table}
\centering
\begin{tabular}{ |p{1.4cm}|p{1.4cm}|p{1.4cm}|p{1.4cm}| }
 \hline
 \textbf{Arch}&\textbf{mIoU}&\textbf{Split}\\
 \hline
 ViT\_{small}  &88\% &$R_a$\\
 ViT\_{base}  &72\% &$R_a$ \\
 ViT\_{tiny}  &66\% &$R_a$ \\
 ResNet50   &36\% &$R_a$ \\
 \hline
\end{tabular}
\caption{Performance of \tseg using different architectures.}
\label{table:arciPerformance}
\end{table}

\begin{table}
\centering
\begin{tabular}{ |p{2.4cm}|p{1.4cm}| }
 \hline
 \textbf{Framework}&\textbf{mIoU}\\
 \hline
 \tseg  &88\% \\
 HRNet  &67\%  \\
 FPN  &87\%  \\
 \hline
\end{tabular}
\caption{Mean-IoUs for \re dataset.}
\label{table:arciPerformance1}
\end{table}
\begin{table}
\centering
\begin{tabular}{ |p{2.4cm}|p{1.4cm}| }
 \hline
 \textbf{Framework}&\textbf{mIoU}\\
 \hline
 \tseg  &89\% \\
 \hline
\end{tabular}
\caption{Mean-IoUs for \ru dataset.}
\label{table:arciPerformance2}
\end{table}

\subsection{Implementation Details}
\subsubsection{Dataset:} 
We evaluate our \tseg on two recent off-road datasets \re and \ru. \re   dataset consists of 20 classes and 6,235 labelled images. The original resolution of the image is $1920~\times~1600$ but for training purposes we downscale the image to $512~\times~512$. With \re dataset we segment unique classes like rubble, mud, and man-made barriers which weren't present in \ru. 

\ru  dataset has 7,456 labelled images, that is for every fifth frame in the sequence. It has 24 different classes including categories like sign, rock and bridge. The original size of the image is $688~\times~550$. 
 Among these thousands of images We use the automatic data selection algorithm to select the most diverse set of images and provide only their annotation for semi-supervised video sequence segmentation in both these datasets.

\subsubsection{Training Details: } We follow DINO \cite{caron2021emerging} and fine tune the network on \re  without labels. The training is completely unsupervised and is performed with AdamW optimizer\cite{loshchilov2018fixing} and 5 batch size using the ViT-8 model. The output patch tokens are then evaluated on two offroad datasets \re and \ru for semantic segmentation. For the semi supervised video sequence segmentation instead of providing label of only the first frame in the sequence, we run ADS for each sequence to select the most diversed images and provide their label. For ADS we follow \cite{hoyer2020three}, it has a slimmed network architecture with a ResNet50 encoder and fewer decoder channels. We iterate the number of selected sample for labelling as 25 and 744. For both subsets, a student depth error is calculated when the student depth network is trained from scratch for 4k and 20k iterations.
\begin{figure*}[h!]
\centering
    \includegraphics[width=11cm, height=5cm]{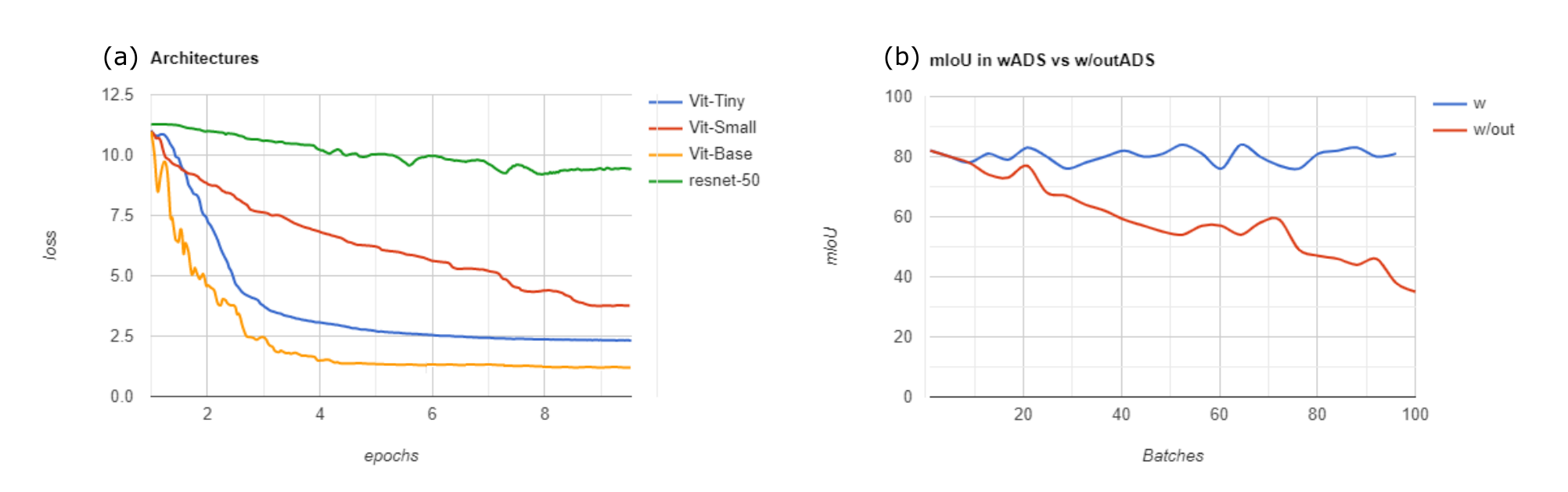}
    \caption{(a) Training parameters for loss degradation of using different backbones in DINO transformer network in \re dataset. (b) Representation of mIoUs obtained from only transformer network and transformer with ADS.}
    \label{fig:graph}
\end{figure*}

\begin{figure*}[h!]
    \centering
    \includegraphics[width=15cm, height=10cm]{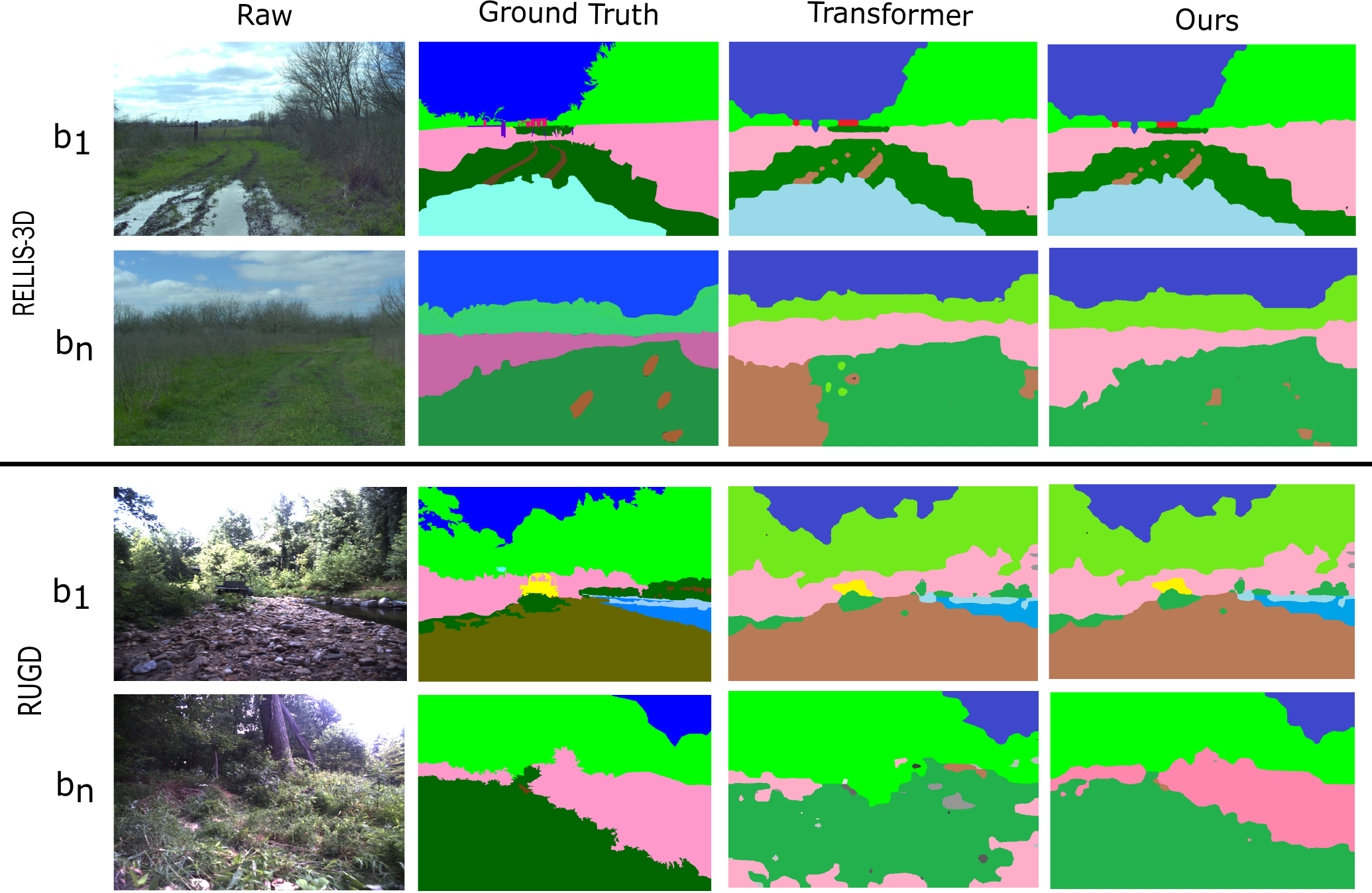}
    \caption{Segmentation results: $b_{1}$ represents the outputs from batch 1 of during semi supervised segmentation. Whereas, $b_{n}$ represents the outputs from the last batch.  }
    \label{fig:datasetresults}
\end{figure*}

\subsection{Ablation studies}
To evaluate the performance of our proposed \tseg approach we calculate the individual class IoU on \re dataset and compare it with the the convolutional network based network HRNet. Table \ref{table:rellis} shows that \tseg solves the class imbalance problem by giving the IoU value of $96\%$ which was $0\%$ in CNN based network HRNet as provided in the \re benchmark. The comparison also shows that with our approach we get state of the art IoU values in majority of the categories. We evaluate \tseg on one another dataset, \ru and the get the values for all different categories as shown in table \ref{table:rugd1} compared with table \ref{table:rugd2}. Figure \ref{fig:labelscompare} shows the visible differences in the performance of different networks. The \re dataset raw image has 7 classes and HRNet is only able to segment 4 of those classes whereas with \tseg we segment all classes including log, rubble and barrier. The mean-IoU(mIoU) value for the two datasets in table \ref{table:arciPerformance1} and table \ref{table:arciPerformance2} shows that using transformer and automatic data selection can give the state of the art values in segmentation of unstructured outdoor environments. We also analyze how different backbones of the self-supervised transformer network impact the segmentation performance in the offroad datasets by using four different architectures $ViT_{small}$, $ViT_{base}$, $ViT_{tiny}$ and $ResNet-50$. We fine-tune off-road \re dataset in an unsupervised fashion using vision transformer architectures and CNN architecture. The performance is shown in table \ref{table:arciPerformance}. Using our data selection approach with four different architectures we see that $ViT_{small}$ gives the best performance with $88\%$ mIoU and on the other hand with ResNet-50 has $36\%$ mIoU value. Figure \ref{fig:graph} summarizes these results with different number of epochs in a form of graph.

We run ADS algorithm on each image sequence and select the best set of 25 images in \re and \ru. For semi-supervised video segmentation instead of providing the label for only the first image in the sequence, we provide label masks for all these 25 images and divide the sequence into 25 batches. Since, these selected images are well diversed with all the features we see significant improvement in segmentation of offroad images. We show the comparison in figure \ref{fig:datasetresults} where $b_{1}$ represents first batch of the sequence and $b_{n}$ represents the last batch. We used the transformer based network DINO to evaluate video segmentation on \re and \ru, however, when the label for only the first image of the sequence is provided we see good performance in $b_{1}$ but the performance is poor as we proceed to the last batch $b_{n}$ of the sequence. It shows that introducing ADS algorithm with transformer based network can significantly improve the segmentation even for the last batch, for example the last batch of the \ru dataset in the figure successfully segments log with grass and bush.


\section{Conclusion}
\label{sec:conclusion}
In this work, we have explored transformer based  approach for semi-supervised segmentation in unstructured outdoor environments along with
handling the class imbalance problem in two most famous offroad datasets. We show how the scene layout information present in the features of vision transformer can be used for semi-supervised segmentation in offroad datasets. Annotation is an important part of the segmentation task in off-trail environments because the performance of the network depends on how precisely the data is labelled. But on the same hand labelling an offroad dataset is particularly challenging because of undefined boundaries. Hence in our approach, we propose to use an Automatic Data Selection algorithm based on depth estimation which selects the most feature rich images. This significantly reduces the number of annotated images needed while training and evaluating with increase in performance. The experiment results show that by introducing \tseg we get state-of-the-art results in unstructured environments. But our approach also has some limitations, it requires a human in loop to annotate the images after getting the most diversed set of images selected for semi-supervised segmentation, for future work the labelling process in between the loop can be automated which will further reduce the chances inaccuracy.

\bibliographystyle{IEEEtran}
\bibliography{bibliography.bib}
\end{document}